\documentclass{article} %
\usepackage{etoolbox}       %

\usepackage{amsmath,amsthm}   %
\usepackage{mathtools}  %
\usepackage[dvipsnames]{xcolor}         %
\usepackage[square,sort,comma,numbers,round]{natbib}
\usepackage{multicol}

\newtoggle{arxiv}
\toggletrue{arxiv}
\iftoggle{arxiv}{
  \usepackage[parfill]{parskip}
  
  \setlength{\textwidth}{6.8in}  %
  \setlength{\textheight}{9in}
  \setlength{\oddsidemargin}{0in}
  \setlength{\evensidemargin}{0in}
  \setlength{\topmargin}{-0.5in}
  \newlength{\defbaselineskip}
  \setlength{\defbaselineskip}{\baselineskip}
  \setlength{\marginparwidth}{0.8in}
  \setlength{\parskip}{6pt}%
  \setlength{\parindent}{0pt}%

  \RequirePackage[T1]{fontenc}
  \RequirePackage[tt=false, type1=true]{libertine}
  \RequirePackage[varqu]{zi4}
  \RequirePackage[libertine]{newtxmath}

}{
  \usepackage{amsfonts}       %
  \addtolength{\tabcolsep}{-3pt} %
  \def\setstretch#1{\renewcommand{\baselinestretch}{#1}}
  \setstretch{0.98}
  \addtolength{\parskip}{-1pt}

  \usepackage[compact]{titlesec}
  \titlespacing{\section}{0pt}{*1}{*0}
  \titlespacing{\subsection}{0pt}{*1}{*0}
  \usepackage[subtle, mathdisplays=tight, charwidths=tight, leading=normal]{savetrees}
  \addtolength\textfloatsep{-0.5em}
  \addtolength\intextsep{-0.2em}
}

\usepackage{bm}
\usepackage{bbm}

\usepackage{booktabs}       %
\usepackage{nicefrac}       %
\usepackage{dblfloatfix}
\usepackage{microtype}      %
\usepackage{multirow}
\usepackage[inline]{enumitem}
\usepackage{diagbox}
\usepackage{comment}
\usepackage{graphicx}
\usepackage{float}
\usepackage{wrapfig}
\usepackage[font=small]{caption} %
\usepackage{algorithm,algorithmicx,algpseudocode} %
\usepackage{subcaption}
\usepackage{colortbl} 
\usepackage{xcolor} 

\usepackage{mathtools}

\DeclarePairedDelimiter\floor{\lfloor}{\rfloor}

\definecolor{mygray1}{gray}{0.95}
\definecolor{mygray}{gray}{0.9}

\usepackage{url}
\usepackage{hyperref}
\hypersetup{
     colorlinks   = true,
     linkcolor    = blue,
     citecolor    = magenta,
     urlcolor     = red,
     pdfborderstyle={/S/U/W 1},
}

\title{\mbox{Kolmogorov–Arnold Transformer}}

\date{} 					

\author{ Xingyi Yang \quad Xinchao Wang \\
	National University of Singapore \\
	\texttt{xyang@u.nus.edu; xinchao@nus.edu.sg} \\
}

\begin{document}
\maketitle
\begin{figure}[h]
    \centering
    \includegraphics[width=0.8\linewidth]{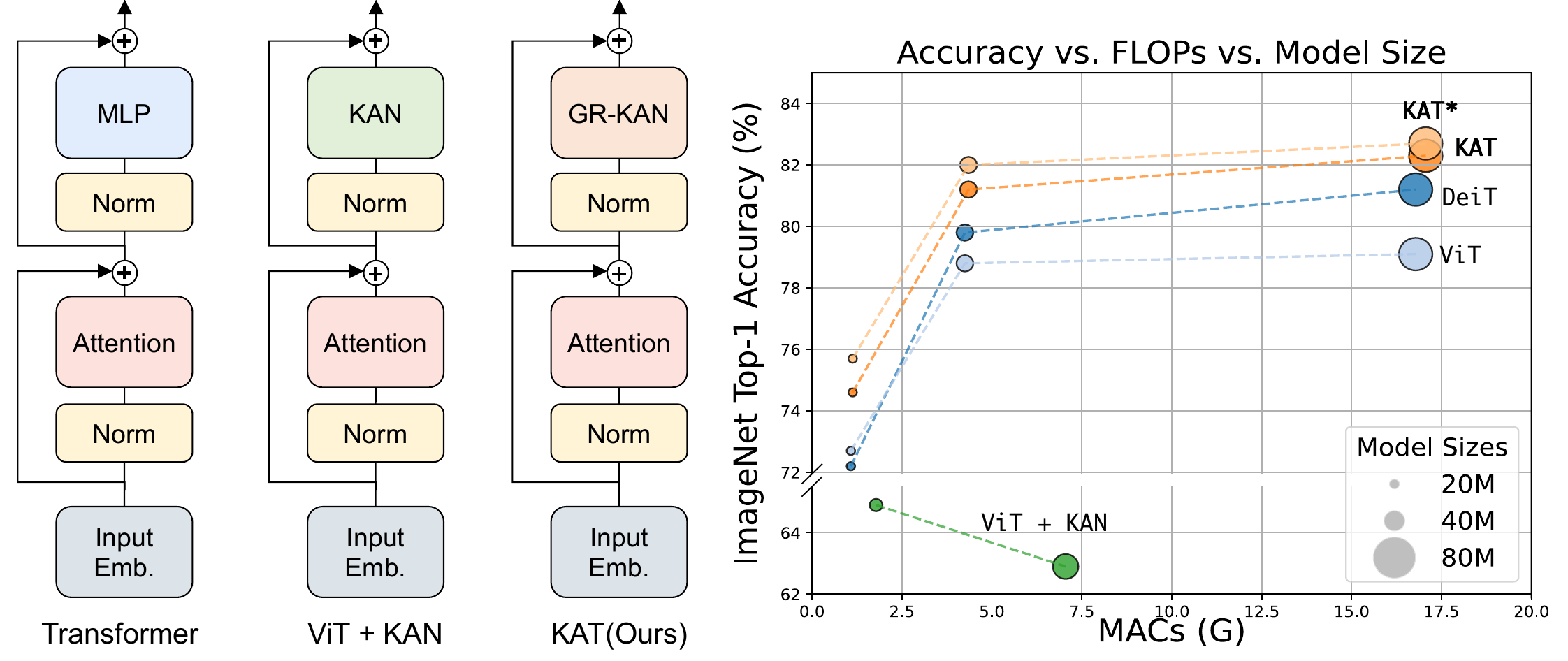}
    \caption{(Left) Architecture of standard transformer (e.g. ViT), ViT+KAN which substitutes the MLP with a KAN, and our KAT model. In KAT, the MLP layers in transformers are replaced with GR-KAN layers. (Right) Performance on the ImageNet dataset. KAT$^*$ indicates that the model was initialized using a pre-trained ViT. Generally, KAT outperforms both the ViT and DeiT models. ViT+KAN performs poorly on ImageNet-level training.}
    \label{fig:kat}
\end{figure}

\begin{abstract}
Transformers stand as the cornerstone of mordern deep learning. Traditionally, these models rely on multi-layer perceptron (MLP) layers to mix the information between channels. In this paper, we introduce the Kolmogorov–Arnold Transformer (KAT), a novel architecture that replaces MLP layers with Kolmogorov-Arnold Network (KAN) layers to enhance the expressiveness and performance of the model. Integrating KANs into transformers, however, is no easy feat, especially when scaled up. Specifically, we identify three key challenges: (C1) \textit{Base function}. The standard B-spline function used in KANs is not optimized for parallel computing on modern hardware, resulting in slower inference speeds. (C2) \textit{Parameter and Computation Inefficiency}. KAN requires a unique function for each input-output pair, making the computation extremely large. (C3) \textit{Weight initialization}. The initialization of weights in KANs is particularly challenging due to their learnable activation functions, which are critical for achieving convergence in deep neural networks. To overcome the aforementioned challenges, we propose three key solutions: (S1) \textit{Rational basis}. We replace B-spline functions with rational functions to improve compatibility with modern GPUs. By implementing this in CUDA, we achieve faster computations. (S2) \textit{Group KAN}. We share the activation weights through a group of neurons, to reduce the computational load without sacrificing performance. (S3) \textit{Variance-preserving initialization}. We carefully initialize the activation weights to make sure that the activation variance is maintained across layers. With these designs, KAT scales effectively and readily outperforms traditional MLP-based transformers. We demonstrate the advantages of KAT across various tasks, including image recognition, object detection, and semantic segmentation. It consistently enhances performance over the standard transformer architectures of different model sizes. Our code is openly available at \url{https://github.com/Adamdad/kat}.
\end{abstract}


\section{Introduction}

Transformers have become the \textit{de facto} architecture in deep learning, widely adopted in computer vision~\cite{dosovitskiy2020vit} and natural language processing~\cite{DBLP:conf/nips/VaswaniSPUJGKP17}. At their core, transformers are built upon two fundamental components: attention modules and multi-layer perceptrons (MLPs). Although significant research has focused on replacing the traditional attention mechanism with alternative operations~\cite{liu2021swin,liu2022convnet,tolstikhin2021mlp}, these variants still lean heavily on MLPs. Surprisingly, there have been relatively few efforts~\cite{shazeer2020glu} aimed at enhancing MLPs themselves. 

Opening up the box, MLPs are composed of stacked linear layers coupled with non-linear activations. What makes it so popular is that, theoretically, they can approximate any function, assuming that there are enough neurons available~\cite{hornik1989multilayer}. 

However, despite their versatility, MLPs face limitations in modeling complex functions. For example, when using ReLU-like activation, a two-layer MLP may struggle to fit periodic functions. Moreover, employing gradient descent to train these networks often results in prolonged convergence times for high-frequency components~\cite{rahaman2019spectral,basri2020frequency,ronen2019convergence}. These challenges have led researchers to explore alternative, perhaps more expressive architectures than MLPs.

Recently, Kolmogorov-Arnold Networks (KANs) emerged as a powerful alternative. KANs are noted for their theoretical parameter efficiency, potentially requiring fewer parameters to model complex functions~\cite{liu2024kan}. They are particularly suitable for mathematical or symbolic regression tasks~\cite{yu2024kan,bozorgasl2024wav,liu2024kan2}. 
The key to such success is the learnable base function in each input-output pair. Those functions are often parameterized by B-spline curves~\cite{unser1993b,gordon1974b}. This design allows KANs to approximate more intricate functions through a summation of spline bases.

Given its potential, integrating KAN layers into transformers~\cite{DBLP:conf/nips/VaswaniSPUJGKP17} becomes an exciting topic. Such integration may boost the expressiveness and efficiency of transformers, enhancing their competitiveness across a wide range of applications.

Unfortunately, this ambition has been met with limited success. In particular, KANs have been reported to be ``10$\times$ slower than MLPs, given the same number of parameters''. Initial attempts to apply KANs to vision recognition tasks have yielded disappointing results. Even on a small scale, these studies have consistently fallen short of matching, let alone surpassing, the performance of traditional architectures. This lack of improvement is often attributed to the limited computational resources and ongoing scalability problems~\cite{cheon2024demonstrating,bodner2024convolutional,cheon2024demonstrating,cheon2024kolmogorov}. 

In a preliminary experiment, we attempted to replace MLP layers in the Vision Transformer (ViT) with KAN layers. It creates a model, which we call ViT+KAN. However, as shown in Figure~\ref{fig:kat} (Right), this straightforward substitution led to significant challenges when performing ImageNet-scale training, resulting in poor performance.
Scalability, therefore, remains a significant obstacle for KAN-based models.

\noindent\textbf{Motivation and Challenges.} Through dedicated analysis, we have identified several key challenges that hinder the effectiveness of KANs in large-scale applications, ultimately limiting their scalability.

\begin{itemize}
    \item \textbf{(C1)} \emph{Base function.} The standard B-spline functions in KANs are not ideal for parallel computing architectures typical of modern GPUs. B-splines require recursive computation, which significantly slows down even the most optimized implementations.
\item \textbf{(C2)} \emph{Parameter and Computation Inefficiency.} Each unique input-output pair in a KAN requires a distinct set of parameters and base functions. This necessity causes an exponential growth in the number of parameters as the network's hidden size increases, resulting in substantial computational overhead and scalability issues.
\item \textbf{(C3)} \emph{Weight initialization.} The weight initialization in KANs is similar to that in MLPs, but it does not meet KANs' needs for convergence. This mismatch can lead to instability and degraded performance during the training process.
\end{itemize}

\noindent\textbf{Our Approach.} In this paper, we introduce \emph{Kolmogorov–Arnold Transformer}~(KAT), which successfully integrates KANs into transformers for large-scale training scenarios such as ImageNet. Beyond simple replacement, We have developed three key innovations~(S1-S3) to address these challenges~(C1-C3) respectively.

\begin{itemize}
    \item \textbf{(S1)} \emph{Rational activation.} We employ rational function as our base function and provide full CUDA implementation.  It aligns better with modern GPU architectures, enhancing computational efficiency and compatibility.
    \item \textbf{(S2)} \emph{Group KAN.} We share function coefficients and base functions among groups of edges. This strategy reduces computational load significantly without sacrificing performance.
    \item \textbf{(S3)} \emph{Variance-preserving initialization}. We carefully initialize weights to maintain consistent variance in activations across the model's layers. This ensures stability during training and improves the model's learning dynamics.
\end{itemize}

By combining all solutions S1-S3, we present a new variant of KAN, called \emph{Group-Rational KAN}~(GR-KAN), to replace the MLP in transformer. We show that GR-KAN is computationally efficient, easy to implement, and can be seamlessly integrated into vision transformers, replacing MLP layers to achieve superior performance. Furthermore, our designs allow KAT to load pre-trained weights from ViT models and continue training to achieve even better results.

We empirically validate KAT across a range of vision tasks, including image recognition, object detection, and semantic segmentation.
The results demonstrate that KAT outperforms traditional MLP-based transformers, achieving enhanced performance with comparable computational requirements. As illustrated in Figure \ref{fig:kat}, KAT-B achieves 82.3\% accuracy on ImageNet-1K, surpassing the ViT model of the same size by 3.1\%. When initialized with pre-trained weights from ViT, the performance further improves to 82.7\%.

The contributions of our paper are threefold. First, we conduct a thorough analysis of the challenges in scaling KAN-based models, particularly focusing on inefficiencies in base functions, parameterization, and weight initialization. Based on this analysis, we propose a set of solutions: rational activation functions tailored for GPU efficiency, Group KAN to reduce computational overhead, and variance-preserving initialization to ensure stable training. Second, leveraging these insights, we introduce the Kolmogorov–Arnold Transformer (KAT) and scale it to ImageNet-level training, successfully integrating KANs into large-scale models. Third, we validate our approach through extensive experiments, showing that KAT not only matches but surpasses the performance of ViT models, all under similar computational requirements.

\section{Preliminary}
\subsection{Kolmogorov-Arnold representation theorem}

The Kolmogorov-Arnold representation theorem~\cite{hecht1987kolmogorov} states that any multivariate continuous function $f$, defined on a bounded domain, can be expressed as a finite composition of continuous univariate functions and addition. Specifically, for a smooth function $f: [0,1]^n \to \mathbb{R}$, it can be represented as:

\[
f(x_1, \ldots, x_n) = \sum_{q=1}^{2n+1} \Phi_q\left(\sum_{p=1}^{n} \phi_{q,p}(x_p)\right)
\]

Here, each function $\phi_{q,p}: [0,1] \to \mathbb{R}$ and $\Phi_q: \mathbb{R} \to \mathbb{R}$ are continuous. This means that the (2d+1)(d+1) univariate functions $\Phi_q$ and $\phi_{q,p}$ are enough for an exact representation of a d-variate function.

This theorm can be written in matrix form as follows:
\begin{equation}
    f(\mathbf{x}) = \Phi_{\text{out}} \circ \Phi_{\text{in}} \circ \mathbf{x} \label{eq:kan_vector}
\end{equation}
where $\Phi_{\text{in}}$ and $\Phi_{\text{out}}$ are defined as:
\begin{equation}
    \Phi_{\text{in}} = \begin{bmatrix}
\phi_{1,1}(\cdot) & \cdots & \phi_{1,n}(\cdot) \\
\vdots & \ddots & \vdots \\
\phi_{2d+1,1}(\cdot) & \cdots & \phi_{2d+1,d}(\cdot)
\end{bmatrix}
\end{equation}
\begin{equation}
    \Phi_{\text{out}} = \begin{bmatrix}
\Phi_1(\cdot) & \cdots & \Phi_{2d+1}(\cdot)
\end{bmatrix}
\end{equation}

This decomposition illustrates how $f$ can be built from simpler functions, showcasing an essential property of multivariate continuous functions.

\subsection{Kolmogorov–Arnold Networks}

Inspired by the Kolmogorov-Arnold representation theorem, \cite{liu2024kan} define a generalized Kolmogorov-Arnold layer to learn univariate functions on edge, in the form of activation function. Formally, a Kolmogorov-Arnold layer with $d_\text{in}$-dimensional inputs and $d_\text{out}$-dimensional outputs is illustrated as
\begin{equation}
    f(\mathbf{x}) = \Phi \circ \mathbf{x} = \begin{bmatrix}\sum_{i=1}^{d_{in}} \phi_{i,1}(x_i) & \dots& \sum_{i=1}^{d_{in}} \phi_{i,d_{out}}(x_i)\end{bmatrix}, \text{where}\quad \Phi = \begin{bmatrix}
\phi_{1,1}(\cdot) & \cdots & \phi_{1,d_{\text{in}}}(\cdot) \\
\vdots & \ddots & \vdots \\
\phi_{d_{\text{out}},1}(\cdot) & \cdots & \phi_{d_{\text{out}},d_{\text{in}}}(\cdot)
\end{bmatrix}\label{eq:kan_layer}
\end{equation}
Note that Eq~\ref{eq:kan_layer} can be seen as a generalized form of Eq~\ref{eq:kan_vector}, such that $\Phi = \Phi_{\text{in}}\circ \Phi_{\text{out}}$. A general KAN network is a stacking of 
$L$ layers: given an input vector $\mathbf{x}_0 \in \mathbb{R}^{d_0}$, the output of KAN is $KAN(\mathbf{x}_0) = \Phi_{L-1} \circ \Phi_{L-2} \dots \circ \Phi_0 \circ \mathbf{x}_0$. 

In practice, \cite{liu2024kan} parameterizes $\Phi$ use a linear combination of SiLU activation~\cite{elfwing2018sigmoid} and a B-spline function
\begin{align}
        \phi(x) = w_b \texttt{silu}(x) + w_s \texttt{spline}(x), &\text{where} \quad\texttt{silu}(x) = \frac{x}{1+e^{-x}}, &\texttt{spline}(x)=\sum_i c_i B_i(x)
\end{align}

\section{Why original KAN fails to scale?}
\label{sec:fail}
This section examines the scalability issues of KAN. We will explore three key factors: the choice of base function, redundant parameters and computation, and initialization problems. These design choices make the vanilla version of KAN resource-intensive and difficult to apply to large-scale models.

\paragraph{B-spline is not GPU Friendly.} The use of B-spline functions in KAN layers introduces challenges when implemented on GPUs. \textbf{First}, B-splines are not standard functions within CUDA. Implementing them using pure PyTorch and NumPy results in slower performance on modern GPU devices due to the lack of optimized CUDA support. \textbf{Second}, the localized nature of B-Spline computations complicates their use in parallel GPU processes. Typically, each control point influences only a small adjacent area of the curve. This leads to sparse or recursive computations, a type of operation that GPUs manage less efficiently. Although there are efficient implementation for cubic B-Spline~\cite{ruijters2012gpu,ruijters2008efficient,sigg2005fast}, scaling these methods to higher orders is not straightforward.

\paragraph{Parameter and Computation Inefficiency.} Unlike standard neural networks, KAN employs a learnable base function for each pair of input-output channels. This design inherently leads to an increased parameter count and higher computational demands, especially when scaling up the width and depth of a neural network.

In the standard configuration of KAN, a layer with $d_in$ input and $d_out$ output channels incorporates an B-spline function for each input-output pair, of order $K$ on $G$ intervals. This results in the network having a total of $(d_{in} \times d_{out})\times (G+K+3) + d_{out}$ learnable parameters. In contrast, a typical MLP only needs $(d_{in} \times d_{out}) + d_{out}$ parameters.

In terms of computation, the FLOPs for one sample\footnote{For full computation derivation, please see~\cite{yu2024kan}.} in B-spline with De Boor-Cox Iterative~\cite{boor1971subroutine} formulation is $\Big\{\text{FLOPs of non-linear function} \times d_{in} +(d_{in} \times d_{out})\times [9 K\times(G+1.5 K)+2G - 2.5 K+3]\Big\}$. Meanwhile, the FLOPs for an equivalent MLP layer is merely $\Big\{\text{FLOPs of non-linear function} \times d_{out} +2\times (d_{in} \times d_{out})\Big\}$. 

Overall, the parameter size and computational effort of KAN are on the order of $O(G+K)$ and $O(GK)$ times greater than those of a conventional MLP, respectively. This significant increase in complexity is a primary reason why KAN struggles to scale effectively.

\noindent\textbf{Weights are not Properly Initialized.} Deep learning heavily relies on good weight initialization to enable trainability and convergence. A fundamental principle is to ensure  \emph{variance-preserving}, meaning that the variance of the signal should remain constant as it propagates through multiple layers, whether forward or backward~\cite{lecun2002efficient,glorot2010understanding,he2015delving}. This principle ensures that the activation and gradient maintain stability across layers.

However, in the KAN paper, the initialization strategy deviates from this principle. Specifically, the B-spline coefficients $c_i$ are initialized as $\mathcal{N}(0, \sigma^2)$ with $\sigma=0.1$, and $w_s=1$ and $w_b\sim U[-\frac{6}{\sqrt{d_{in} + d_{out}}}, \frac{6}{\sqrt{d_{in} + d_{out}}}]$ are initialized according to the Xavier initialization~\cite{glorot2010understanding}. The combined output variance of the model can be expressed as:
\begin{equation}
    Var[\phi(x)] = Var[w_b \texttt{silu}(x)] + Var[w_s \texttt{spline}(x)] = 3 \mathbb{E}[\texttt{silu}^2(x)] + \mathbb{E}[\texttt{spline}^2(x)]
\end{equation}
If we assume the input 
$x$ is normally distributed, $x\sim \mathcal{N}(0, \sigma_x^2)$ and consider a zero-th order spline, the variance of $\texttt{spline}(x)$ at any point $x$ is simply:
\begin{equation}
    \mathbb{E}[\texttt{spline}^2(x)]=\sum_i c_i^2 Var[B_i(x)] = \sigma^2\sum_i Var[B_i(x)] =\sigma^2 = 0.01
\end{equation}
For the SiLU activation function, although exact variance calculations are complex, numerical estimations indicates $\mathbb{E}[\texttt{silu}^2(x)]\approx 0.355 \sigma_x^2 $. Combining these, we find $Var[\phi(x)] \approx 0.01 + 1.064\sigma_x^2 \neq Var[x]$.

This indicates that, under zero-th order spline, $Var[\phi(x)] \neq Var[x]$. With higher-order splines, the variance instability might increase. Thus, the default initialization opposes the essential variance-preserving principle.

\section{Kolmogorov–Arnold Transformer}
As discussed earlier, the standard KAN faces three major challenges that limit its use in large, deep neural networks. In this section, we refine its design to better suit modern transformers, allowing us to replace MLP layers with KANs.

\subsection{Overall Architecture}
Just as its name imply, Kolmogorov–Arnold Transformer~(KAT) replaces the MLPs in vision transformer~\cite{dosovitskiy2020vit} with KAN layers. 

Specifically, for a 2D image \(\mathbf{x} \in \mathbb{R}^{H \times W \times C}\), we first flatten it into a 1D sequence, apply patch embedding and positional encoding, and then pass it through a series of KAT layers. At layer \(\ell\), the following operations are performed:
\begin{equation}
    \mathbf{x}_0^{(\ell)} = \text{MSA}(\text{LN}(x_{\ell-1})) + \mathbf{x}_{\ell-1}, \quad \ell = 1, \dots, L
\end{equation}
\begin{multicols}{2}
\noindent
\begin{equation}
    \mathbf{x}_{\ell} = \text{MLP}(\text{LN}(\mathbf{x}_0^{(\ell)})) + \mathbf{x}_0^{(\ell)}, \quad [\textbf{Transformer}]
\end{equation}
\columnbreak
\noindent
\begin{equation}
    \mathbf{x}_{\ell} = \text{KAN}(\text{LN}(\mathbf{x}_0^{(\ell)})) + \mathbf{x}_0^{(\ell)}, \quad [\textbf{KAT}] 
    \label{eq:kanreaplcemlp}
\end{equation}
\end{multicols}
where $\mathbf{x}_{\ell}$ stands for the output feature sequence at the $\ell$ layer.
As illustrated, we replace all two-layer MLPs with two-layer KANs while keeping the attention layers unchanged. Although similar efforts have been made in specific  domains~\cite{VisionKAN2024,chen2024sckansformer}, a simple replacement is not enough to achieve scalability in large models. 

Most importantly, here, we introduce a special kind \emph{Group-Rational KAN}. We use rational functions as the base function for KAN~(Section~\ref{sec:rat}) and share parameters between a group of edges~(Section~\ref{sec:group}). We also specify the weight initialization scheme to ensure stable training~(Section~\ref{sec:init}). Together, these enhancements make KAT more scalable and improve performance.

\subsection{Rational Base Functions}
\label{sec:rat}

In our method, we use the rational function~\cite{boulle2020rational,telgarsky2017neural,leung1993rational,aghaei2024rkan} as the base function for the KAN layer, instead of the B-spline. 

Specifically, we parameterize the function $\phi(x)$ on each edge as rational over polynomials $P(x), Q(x)$ of order $m,n$. 
\begin{equation}
    \phi(x) = wF(x) = w\frac{P(x)}{Q(x)} = w\frac{a_0 + a_1x+\dots + a_m x^m}{b_0 + b_1x+\dots + b_n x^n} 
\end{equation}
$a_n$ and $b_m$ are coefficient of the rational function and $w$ is the scaling  factor. This function is said to have degree $m/n$. We hope to learn those $a_n,b_m$ and $w$ through end-to-end backpropagation.

To avoid instability caused by poles, where $Q(x)\to 0$ and $\phi(x)\to \pm\infty$, we employ a Safe Padé Activation Unit (PAU)~\cite{Molina2020Padé} as our basis, which is a modified form of the standard rational function
\begin{equation}
     F(x) = \frac{a_0 + a_1x+\dots + a_m x^m}{1 + |b_1x+\dots + b_n x^n|}\label{eq:rational}
\end{equation}

\noindent\textbf{Why use Rational Function?} There are practical and theoretical reasons for selecting rational functions as our base functions.

First, from an efficiency perspective, evaluating polynomials involves simple operations that are highly suitable for parallel computing. This makes rational functions computationally efficient for large-scale models.

Second, from a theoretical perspective, rational functions can approximate a wider range of functions—including those with singularities or sharp variations—more efficiently and accurately than polynomials~\cite{walsh1935interpolation,baker1961pade}. Since B-splines are essentially sums of local polynomials, rational functions offer a theoretical advantage over B-splines for modeling complex behaviors.

Third, from a practical perspective, rational activations have already been successfully used as activation functions in neural networks~\cite{boulle2020rational,Molina2020Padé}.

Given these reasons, we adopt rational functions as the base functions in our KAN layers to enhance the model's expressiveness, stability, and computational efficiency.

\noindent\textbf{Implement Rational Function on GPU.} With the rational function, a core contribution in this paper is to implement it efficiently on parralized devices like GPU. In stead of using \texttt{pytorch} with automatic differentiation, we implement it fully with CUDA~\cite{nickolls2008scalable}. 
\begin{itemize}
    \item Similar to~\cite{Molina2020Padé}, we compute the explicit gradients of $\frac{\delta F}{\delta a_m},\frac{\delta F}{\delta b_n}$ and $\frac{\delta F}{\delta x}$
    \begin{equation}
        \frac{\delta F}{\delta a_m}=\frac{x^m}{Q(x)}, \quad\frac{\delta F}{\delta b_n} = ax^n \frac{A(x)}{|A(x)|}\frac{P(x)}{Q(x)^2}, \quad\text{and} \quad  \frac{\delta F}{\delta x} = \frac{\delta P(x)}{\delta x} \frac{1}{Q(x)} - \frac{\delta Q(x)}{\delta x}\frac{P(x)}{Q^2(x)}
    \end{equation}
    where $A(x)= b_1x+\dots + b_n x^n$, $\frac{\delta P(x)}{\delta x} = a_1 + 2a_2x + ma_mx^{m-1}$ and $\frac{\delta Q(x)}{x} = \frac{A(x)}{|A(x)|} (b_1 + 2b_2x + nb_nx^{n-1})$. 
    \item To optimize the evaluation of polynomials, we employ Horner's method~\cite{horner1815new}, which reformulates a polynomial in a nested form to reduce the computation:
    \begin{equation}
        a_0 + a_1x+\dots + a_m x^m = a_0+ x(a_1 + x(a_2 + x(\dots)))
    \end{equation}
    This allows the evaluation of a polynomial of degree n with only $n$ multiplications and $n$ additions. By default, we use $m=5$ and $n=4$.
\end{itemize}

Through this efficient CUDA implementation, we largely reduce the computation for each evaluation of the base function. As shown in Table~\ref{tab:base_func}, with a scalar input, the rational function with the Horner method is much cheaper than the B-spline used in the KAN paper.

\begin{table}[h]
    \centering
    \caption{Comparison of FLOPs for different functions. Compared to B-spline function. using Horner's method with the Rational function reduces FLOPs by approximately $9.3\times$ compared to the B-Spline function.}
    \label{tab:base_func}
    \begin{tabular}{l|c}
    \toprule
        Name & FLOPs \\
        \midrule
        B-Spline (G=3, K=3) & 204 \\
        Rational (m=5, n=4) & 46 \\
        Rational (m=5, n=4) w Horner& 21 \\
        \bottomrule
    \end{tabular}

\end{table}

\subsection{Group KAN}
\label{sec:group} 

Instead of learning a unique base function for each input-output pair, we can share their parameters within a group of edges. It reduces the number of parameters and computation. This kind of parameter sharing~\cite{lecun1995convolutional,lecun1989backpropagation} and group-wise computation~\cite{DBLP:conf/nips/VaswaniSPUJGKP17,wu2018group} have been key techniques in neural network design.

\begin{figure}[t]
    \centering
\includegraphics[width=0.8\linewidth]{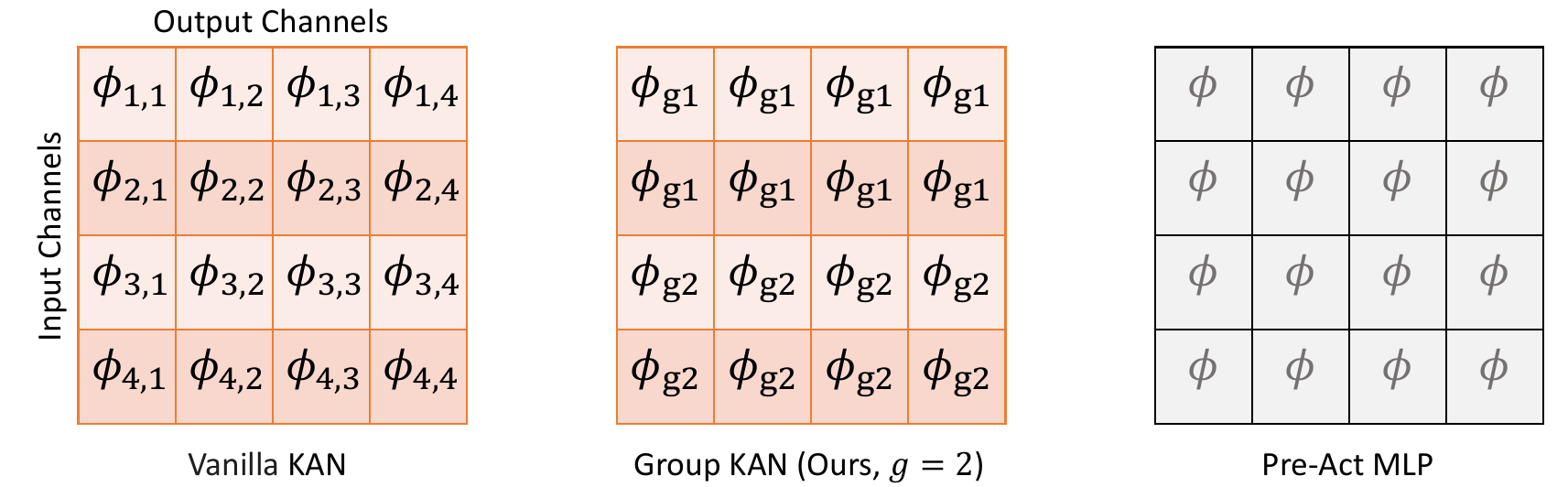}
    \caption{Comparing our Group KAN with vanilla KAN and MLPs. While KAN has unique function on each input-output pairs, Group KAN share these functions at with a groups of edges. }
    \label{fig:groupKAN}
\end{figure}

Specifically, we divide the input channels \(d_{in}\) into \(g\) groups, sharing parameters among \(d_{in}/g\) input channels within each group. Figure~\ref{fig:groupKAN} illustrates the distinctions between the original KAN, our Group KAN, and a standard MLP. Unlike MLPs, which employ non-learnable activations, KAN assigns a unique function to each input-output pair. Group KAN reduces the number of parameters by sharing these functions among a group of edges.

\textbf{Group-Rational KAN.} We combine the rational function of Section~\ref{sec:rat}
 with group-wise parameters to implement our Group-Rational KAN~(GR-KAN). In practice, we share the parameter for the rational function $F$ for each group; however, each edge retains a unique scalar $w$. 

Suppose $i$ is the index of the input channel. With $g$ groups, each group contains $d_g = d_{in}/g$ channels, where $\floor{i/d_g}$ is the group index. The operation of GR-KAN on input vector $\mathbf{x}$ can be expressed as
\begin{align}
        \texttt{GR-KAN}(\mathbf{x}) = \Phi \circ \mathbf{x}  = \begin{bmatrix} \sum_{i=1}^{d_{in}} w_{i,1} F_{\floor{i/d_g}}(x_i) & \dots& \sum_{i=1}^{d_{in}} w_{i,d_{out}} F_{\floor{i/d_g}}(x_i)\end{bmatrix} 
\end{align}
With a simple rewrite, this can be expressed in matrix form as the product of a weight matrix \( \mathbf{W} \in \mathbb{R}^{d_{\text{in}} \times d_{\text{out}}} \) and a input-wise rational function $\mathbf{F}$
\begin{equation}
    \texttt{GR-KAN}(\mathbf{x}) = \mathbf{W}\mathbf{F}(\mathbf{x}) = \begin{bmatrix}
w_{1,1} & \cdots & w_{1,d_{\text{in}}} \\
\vdots & \ddots & \vdots \\
w_{d_{\text{out}},1} & \cdots & w_{d_{\text{out}},d_{\text{in}}} 
\end{bmatrix}\times \begin{bmatrix}  F_{\floor{1/d_g}}(x_1) & \dots& F_{\floor{d_{in}/d_g}}(x_{d_{in}}) \end{bmatrix}^\top 
\end{equation}
As such, we can implement this GR-KAN layer as a group-wise rational function $\mathbf{F}$ followed by a linear layer
\begin{equation}
    \texttt{GR-KAN}(\mathbf{x}) = \texttt{linear}(\texttt{group\_rational}(\mathbf{x}))\label{eq:impl}
\end{equation}

In this form,
sharing parameters across each input channel allows direct application of the rational function to the input vector, equivalently applying it across each grouped edge. In this way, GR-KAN functions as a specialized MLP, with 1) learnable non-linear functions, 2) activation preceding the linear layer, and 3) unique activation functions tailored for each group of edges.

\begin{table}[t]
    \centering
    \begin{tabular}{l|l|l}
    \toprule
        Name & No. Params &FLOPs\\
        \midrule
        MLP & $d_{in} \times d_{out} + d_{out}$ &  $\text{Func FLOPs}\times d_{out}+ 2\times(d_{in} \times d_{out})$\\
        KAN & $d_{in} \times d_{out} \times (G+K+3) + d_{out}$ & $\text{Func FLOPs}\times d_{in} +(d_{in} \times d_{out})\times [9 K\times(G+1.5 K)+2G - 2.5 K+3]$\\
        GR-KAN~(Ours)& $d_{in} \times d_{out} + d_{out} + (m +n \times g )$ & $(2m+2n+3)\times d_{in}+ 2\times(d_{in} \times d_{out})$\\
        \bottomrule
    \end{tabular}
    \caption{Comparison of parameter counts among different models. \emph{Func FLOPs} stands for the computation of used non-linear activation. In KAN, $K$ represents the order number and $G$ the grid number. For our GR-KAN, $m$ and $n$ indicate the order of polynomials, and $g$ represents the number of groups. Our model, GR-KAN, has a parameter size comparable to a constant increase over the MLP, whereas the KAN model's parameters scale with $(G + K + 3)$.}
    \label{tab:param_compute}
\end{table}

In experiments, we notice that for rational function, we share the denominator coefficient $b_n$ among all groups and use different $a_m$ for each group. It gets better performance.

\textbf{Parameter and Computation Savings.} The original KAN requires \(d_{in} \times d_{out}\) unique activation functions. Through our grouping strategy, only \(g\) unique functions are needed, reducing the parameter count to a constant overhead compared to a standard MLP.

Except the saving on parameter number, this grouping also reduces computational demands. Each input channel computes the activation function \( \phi \) once, shared across all corresponding output channels. In contrast, the original KAN requires that each output channel \(j\) to independently compute \( \phi_{i,j} \). This results in significant computational savings. The comparison of the number of parameters and computation is listed in Table~\ref{tab:param_compute}.

\subsection{Variance-Preserving Initialization}
\label{sec:init}
In this section, we aim to initialize the values for $a_m,b_n$ and $w$ in Group-Rational KAN to ensure variance-preserving behavior across the network. At its core, we prevent the growth or reduction of activation magnitudes throughout the layers, thereby maintaining stability.

\begin{figure}[]
    \centering
    \begin{minipage}[b]{0.67\linewidth}
        \centering
        \includegraphics[width=\linewidth]{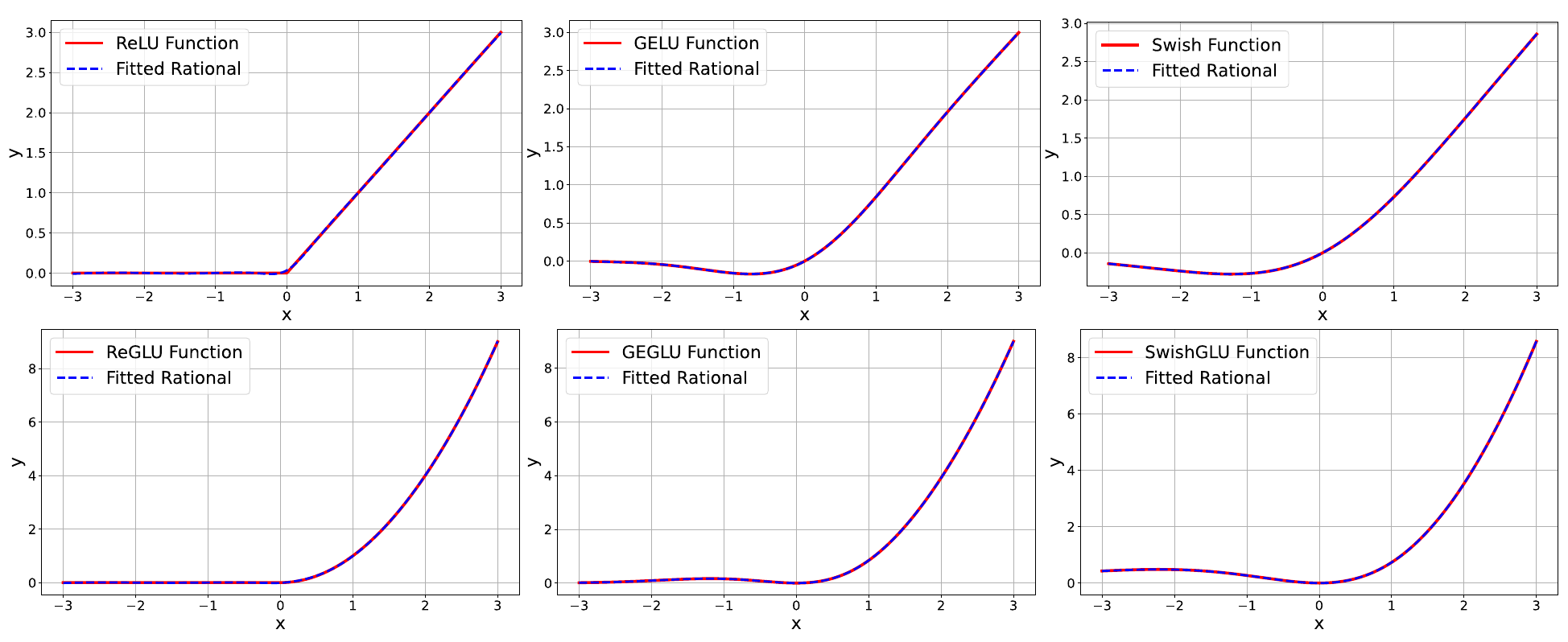}
        \captionof{figure}{Example of fitted functions with rational form.}
        \label{fig:functionfit}
    \end{minipage}
    \hfill 
    \begin{minipage}[b]{0.3\linewidth} 
        \centering
        \begin{tabular}{l|c}
            \toprule
            Name & $\alpha = \frac{Var[x]}{\mathbb{E}[F(x)^2]}$ \\
            \midrule
            Identity & 1 \\
            ReLU & 2 \\
            GELU & 2.3568 \\
            Swish/SiLU & 2.8178 \\
            GEGLU & 0.7112 \\
            SwishGLU & 0.8434\\
            \bottomrule
        \end{tabular}
        \captionof{table}{Expected values of $F(x)^2$ for various functions.}
        \label{tab:gain}
    \end{minipage}%

\end{figure}

We revisit the analysis from~\cite{he2015delving} and adapt it to KANs. For a GR-KAN layer, the computation for each output $y_j$ is given by $y_j = \sum_{i=1}^{d_{in}}\phi(x_i)= \sum_{i=1}^{d_{in}}(w_{i,j} F(x_i)) + b_j$. We assume that all $x_i$ are mutually independent~\cite{glorot2010understanding} and uniformly distributed. Here, $w_{i,j}$ follows a normal distribution $\mathcal{N}(0, \sigma_w^2)$ and $b_j$ is initialized to zero. The variance of $y_j$ can then be described as:
\begin{align}
    Var[y] = d_{in} Var[w F(x)]\\
    Var[y] = d_{in} Var[w] \mathbb{E}[F(x)^2]
\end{align}
where $x$, $y$, and $w$ represent the random variables of each element in $x_i$, $y_j$, and $w_{i,j}$ respectively.
When layers are stacked, we aim for the variance of the input-output activations to remain consistent, expressed as:
\begin{equation}
    Var[x] = d_{in} Var[w] \mathbb{E}[F(x)^2]
\end{equation}
Since $F(x)$ is the rational function containing coefficients $a_m$ and $b_n$, the initialization of $w$ and these coefficients are interdependent—the form of $F(x)$ influences the appropriate initialization of $w$.  The crucial step is to calculate $\frac{Var[x]}{\mathbb{E}[F(x)^2]}$ and adjust $w$ to maintain consistent activation scaling.

For our rational function defined in Equation~\ref{eq:rational}, computing $\mathbb{E}[F(x)^2]$ involves evaluating:
\begin{align}
    \mathbb{E}[F(x)^2] = \int_{-\infty}^{+\infty} F^2(x) f(x)dx = \int_{-\infty}^{+\infty} (\frac{a_0 + a_1x+\dots + a_m x^m}{1 + |b_1x+\dots + b_n x^n|})^2 f(x) dx
\end{align}
where $f(x)$ is the density function of $x$. Unlike activation functions such as ReLU, for which $\mathbb{E}[F(x)^2] = \frac{1}{2} Var[x]$, computing $\mathbb{E}[F(x)^2]$ for the rational function is challenging due to the lack of a closed-form solution. 

\textbf{Initialize $a,b$ first, then initialize $w$.} To make the process manageable, Instead of sampling $w$, $a$, and $b$ jointly, we proceed sequentially. Initially, we determine $a$ and $b$ such that $F$ fits established activations like ReLU, GELU, and Swish. Figure~\ref{fig:functionfit} illustrates the fitted functions.

Once $a$ and $b$ are set, we estimate the gain $\alpha=\frac{\mathbb{E}[F(x)^2]}{Var[x]}$ numerically, assuming $x \sim \mathcal{N}(0,1)$\footnote{This assumption is justified as the inputs to the KAN layer are normalized using layer normalization as in Equation~\ref{eq:kanreaplcemlp}. The LN layers are initialized to have zero bias and a scaling factor of one}.  The calculated gains, $\alpha$, are documented in Table~\ref{tab:gain}. We use the gain value to initialize $w$ from $\mathcal{N}(0,\frac{\alpha}{d_{in}})$.

\textbf{Initialize KAT from ViT.} In addition to random weight initialization, we can also transfer weights from a pre-trained ViT to our KAT model. This transfer is straightforward for most layers, as KAT can replicate the micro-architecture of ViT, except for the KAN layer. 

\begin{wrapfigure}{r}{0.5\textwidth}
\vspace{-6mm}
  \begin{center}
    \includegraphics[width=0.48\textwidth]{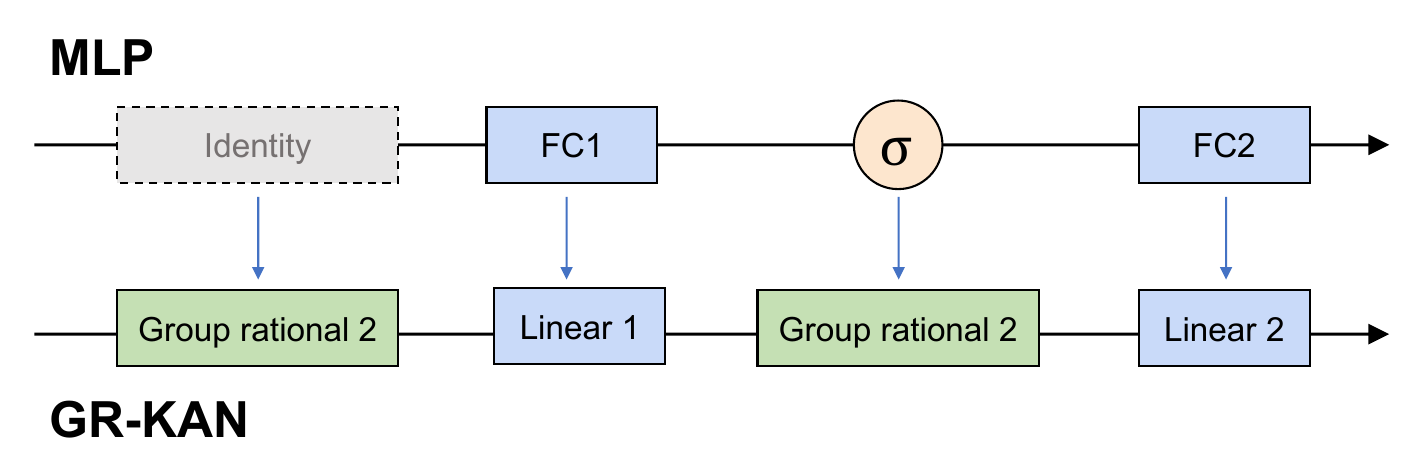}
  \end{center}
  \vspace{-2mm}
  \caption{One-to-one weight mapping between trained MLP in ViT and GR-KAN in KAT.}
  \label{fig:load}
   \vspace{-2mm}
\end{wrapfigure}
For the GR-KAN layer, weight transfer is still feasible, as shown in Figure~\ref{fig:load}. Because the GR-KAN layer consists of a linear layer and a group-wise rational layer, we can directly load the weights of the linear layer from the MLP in the trained ViT. 

For rational layers, the first one is initialized to behave like an identity function, while the second layer is set to approximate the non-linear function used in the original MLP. This approach allows all the weights of the GR-KAN layer to be cloned from a ViT model, ensuring compatibility and efficient initialization.



\section{Experiments}

\subsection{Experimental Setup}
We modify the original ViT~\cite{dosovitskiy2020vit} architecture by substituting its MLP layers with GR-KAN layers. By default, these KAN layers employ a rational function with parameters \(m=5\) and \(n=4\), and are organized into groups of 8~(\(g=8\)). Each transformer block contains 2 KAN layers. The first GR-KAN layer's $a_m$ and $b_n$ are initialized to fit the identity function, while the second is initialized to mimic the Swish function~\cite{ramachandran2017searching}. The attention layers are initialized with Mimetic Initialization~\cite{trockman2023mimetic}. The remainder of the architecture remains unchanged. We intentionally do not use hierarchical architectures~\cite{yu2022metaformer} for simplicity.

\textbf{Model Variant.} We select the configurations of KAT to be identical with those used in ViT~\cite{dosovitskiy2020vit}, as
summarized in Table~\ref{tab:kat_models}. All variants use an input patch size of $16 \times 16$.

\begin{table}[h]
\centering
\begin{tabular}{l|c|c|c|c|c}
\toprule
\textbf{Model} & \textbf{Layers} & \textbf{Hidden Size D} & \textbf{MLP Size} & \textbf{Heads} & \textbf{Params} \\ \midrule
KAT-Tiny  & 12  & 192 & 768  & 3  & 5.7M   \\ 
KAT-Small & 12 & 384 & 1536 & 6  & 22.1M  \\ 
KAT-Base  & 12 & 768 & 3072 & 12 & 86.6M  \\ \bottomrule
\end{tabular}
\caption{Details of KAT model variants.}
\label{tab:kat_models}
\end{table}

\subsection{Image Recognition}

\textbf{Experiment Setup.} We do experiments on ImageNet-1K [59] image classification benchmark. ImageNet-1K is one of the most
widely-used datasets in computer vision which contains
about 1.3M images of 1K classes on training set, and 50K images on validation set.

We mainly follow the hyper-parameters of DeiT~\cite{touvron2021training}. Specifically, models are trained for 300 epochs at  $224^2$
resolution. The patch size is set to 16. Data augmentation and regularization techniques include RandAugment~\cite{cubuk2020randaugment}, Mixup~\cite{DBLP:conf/iclr/ZhangCDL18}, CutMix~\cite{yun2019cutmix}, Random Erasing~\cite{zhong2020random}, weight
decay, Label Smoothing~\cite{szegedy2016rethinking} and Stochastic Depth~\cite{huang2016deep}. We adopt
AdamW~\cite{loshchilov2019decoupledweightdecayregularization} optimizer with batch size of 1024. 

We compare with ViT~\cite{dosovitskiy2020vit} and DeiT~\cite{touvron2021training}, as we share the same architecture, except for the channel mixer. We also report the results of ViT + KAN~\cite{VisionKAN2024}, that simply replacing MLP with standard KAN.

\begin{table}[h]
    \centering
        \caption{Comparative Analysis of Model Performance and Computational Efficiency on ImageNet-1K. We measure the FLOPs under $224^2$ using \texttt{fvcore} package. $*$ indicates that the model is initialized using a pre-trained ViT model, otherwise trained from scratch.}
    \label{tab:imagenet}
    \begin{tabular}{l| c |c|c|c}
    \toprule
        \textbf{Model} & \textbf{Channel Mixer} & \textbf{\#Param.} & \textbf{FLOPs} &\textbf{IN-1k Top-1}  \\
        \midrule
        ViT-Ti/16 & MLP& 5.7M & 1.08G & 72.7 \\
        DeiT-T & MLP & 5.7M & 1.08G &72.2 \\
        ViT-T + KAN & KAN & 12.8M & 1.78G &64.9 \\
        \rowcolor{mygray1}KAT-T & KAN& 5.7M 
        &1.13G &\textbf{74.6} \\
        \rowcolor{mygray}KAT-T$^*$ & KAN& 5.7M &1.13G &\textbf{75.7} \\
        \midrule
        ViT-S/16 & MLP & 22.1M & 4.25G & 78.8\\
        DeiT-S & MLP&22.1M & 4.25G  & 79.8 \\
        ViT-S + KAN  &KAN & 50.4M & 7.05G & 62.9\\
        \rowcolor{mygray1} KAT-S & KAN & 22.1M & 4.35G& \textbf{81.2} \\
        \rowcolor{mygray} KAT-S$^*$ & KAN & 22.1M & 4.35G& \textbf{82.0} \\
        \midrule
        ViT-B/16 & MLP& 86.6M & 16.87G & 79.1\\
        DeiT-B & MLP& 86.6M & 16.87G & 81.8\\
       ViT-B + KAN & KAN & 199.8M &  28.04G & \texttt{NAN} \\
        \rowcolor{mygray1} KAT-B & KAN& 86.6M & 17.06G & \textbf{82.3} \\
         \rowcolor{mygray} KAT-B$^*$ & KAN& 86.6M & 17.06G & \textbf{82.8} \\\bottomrule
    \end{tabular}
\end{table}

\textbf{Results.} Our experimental results demonstrate that the KAT models consistently outperform their counterparts on the IN-1k dataset, as shown in Table~\ref{tab:imagenet}. Firstly, the integration of GR-KAN in the transformer architectures demonstrates superior performance over traditional MLP-based mixers. For instance, the KAT-S model achieved an accuracy of 81.2\%, outperforming the DeiT-S model by 2.4\%. This improvement underscores the potential of KAN mixers to enhance model efficacy when properly integrated.

Secondly, the vanilla KAN layer faces scalability issues. ViT-T/S + KAN only achieved an accuracy of around 63\%, even with a much higher computational cost. ViT-L + KAN fails to converge, resulting in \texttt{NAN} error. We addressed these scaling challenges as detailed in Section~\ref{sec:fail}, enabling our KAT models to scale successfully.

These findings highlight the efficacy of the KAT approach in balancing computational efficiency with improved performance, suggesting valuable directions for further research in optimizing transformer architectures.

\subsection{Object Detection and Instance Segmentation}

\textbf{Experimental Setup.} We evaluate our approach on the MS-COCO2017~\cite{lin2014microsoft} dataset, a standard benchmark for object detection and instance segmentation. In our setup, the KAT is employed as the backbone within a ViTDet-based~\cite{li2022exploring} Mask R-CNN~\cite{he2017mask} model, initialized with weights pre-trained on ImageNet. We followed the standard $3\times$ training schedule, which consists of 36 epochs. The training images were resized to $800\times1333$ pixels. The AdamW optimizer~\cite{loshchilov2019decoupledweightdecayregularization} was used with a learning rate of 0.0001 and a total batch size of 16. Our implementation was based on the PyTorch and MMDetection~\cite{chen2019mmdetection} libraries, and we use FP16 precision to reduce training costs. The experiments were carried out on 4 NVIDIA H100 GPUs.

\textbf{Results.} Table~\ref{tab:mask_rcnn_performance} compares the performance of different backbones. KAT consistently outperformed other models, particularly in object detection, where it achieved a 3.0 AP$^{\text{box}}$ gain on the S-sized model and a 1.4 AP$^{\text{box}}$ gain on the L-sized model compared to ViTDet. The improvements were most pronounced in smaller models, where computational cost increased by only 1 GFLOPs. This shows that KAT offers better accuracy with minimal overhead.

\begin{table}[htbp]
\centering
\caption{Performance of Mask-RCNN with different backbones on 3$\times$ schedule.}
\label{tab:mask_rcnn_performance}
\begin{tabular}{lcccccccc}
\toprule
\textbf{Backbone} &  \textbf{\#Param.} &\textbf{FLOPs} & AP$^{\text{box}}$ & AP$^{\text{box}}_{50}$ & AP$^{\text{box}}_{75}$ & AP$^{\text{mask}}$ & AP$^{\text{mask}}_{50}$& AP$^{\text{mask}}_{75}$ \\
\midrule
PVT-Small & 44.1M & - & 43.0& 65.3 & 46.9 & 39.9 & 62.5 & 42.8 \\ 
Swin-T & 48M &267G &  46.0 & 68.1 & 50.3 & 41.6 & 65.1 & \textbf{44.9} \\
ConvNeXt-T  & 48M& 262G &  46.2 & 67.9 & 50.8 & 41.7 & 65.0 & \textbf{44.9} \\
ViT-S &43.8M &423G & 44.0 & 66.9 & 47.8 & 39.9 &  63.4  &42.2 \\
ViTDet-S & 44.5M & 423G  & 44.5 & 66.9 & 48.4 & 40.1 & 63.6 & 42.5\\
\rowcolor{mygray}KATDet-S & 44.5M&  424G & \textbf{47.5} & \textbf{69.0} &\textbf{51.2}& \textbf{41.5} &\textbf{65.7} & 44.0\\
\midrule
ViT-B  & 113.6M & 767G &45.8 & 68.2 & 50.1 & 41.3 & 65.1& 44.4\\
ViTDet-B & 113.6M & 767G & 46.3 & 68.6 & 50.5 & 41.6 & 65.3 & \textbf{44.5 }\\
\rowcolor{mygray}KATDet-B & 113.7M & 770G & \textbf{47.7} & \textbf{69.1} & \textbf{51.6}& \textbf{41.6} & \textbf{65.9} & 44.3\\
\bottomrule
\end{tabular}
\end{table}

\subsection{Semantic Segmentation}

\textbf{Experiment Setup.} We evaluated our KAT model on the ADE20K dataset~\cite{zhou2017scene}. This dataset comprises 150 semantic categories with 20,000 images in the training set and 2,000 in the validation set. For our experiments, we utilized KAT as the backbone for the UperNet framework~\cite{xiao2018unified}, initializing it with ImageNet pre-trained weights. The training was conducted using the AdamW optimizer~\cite{loshchilov2019decoupledweightdecayregularization} with a learning rate of 0.0001 and a batch size of 16, across 160,000 iterations. Our implementation was carried out using the PyTorch and mmsegmentation libraries, and the experiments were performed on two NVIDIA H100 GPUs. For comparison, we evaluated UperNet with other backbones, including DeiT, Swin Transformer, and ConvNeXt.

\begin{table}[htbp]
\centering

\begin{minipage}{0.48\linewidth}
\caption{Performance of Semantic segmentation with UperNet on ADE20K validation set. Images are cropped to $512\times 512$ for training. The MACs are measured with input size of $512\times 2048$.}
\label{tab:semantic_segmentation}
\centering
\begin{tabular}{l|c|c|c}
\toprule
\textbf{Backbone} & \textbf{\#Param.} & \textbf{FLOPs} & \textbf{mIoU (\%)} \\ \midrule
Swin-T & 60M & 945G & 45.8 \\ 
ConvNeXt-T & 60M & 939G & 46.7 \\ 
DeiT-S & 57M & 1217G & 43.5\\
\rowcolor{mygray} KAT-S & 57M & 1219G & 46.1\\ 
\midrule
Swin-B &  121M & 1188G & 49.5 \\ 
ConvNeXt-B & 122M & 1170G & 49.6 \\ 
DeiT-B & 142M & 2007G & 47.2\\
\rowcolor{mygray} KAT-B & 142M & 2011G & 47.4 \\ 
\bottomrule
\end{tabular}
\end{minipage}%
\hfill
\begin{minipage}{0.48\linewidth}
\centering
\caption{Ablation on activation function, with ViT-Ti/16 Variant.}
\label{tab:act_acc}
\begin{tabular}{l|c|c}
\toprule
\textbf{Name} & \textbf{Learnable?} & \textbf{IN-1k Top-1} \\ \midrule
GELU (Default) & No & 72.7 \\
ReLU & No & 72.8 \\
SiLU & No & 69.8 \\
PReLU & Yes & 73.2 \\
PAU & Yes & 73.6 \\
\rowcolor{mygray} KAT-T & Yes & \textbf{74.6} \\
\bottomrule
\end{tabular}
\end{minipage}
\end{table}





\textbf{Results.} Table~\ref{tab:semantic_segmentation} summarizes the segmentation results. Overall, KAT demonstrates a competitive improvement over plain ViT-based architectures, achieving a 2.4\% improvement over DeiT-S and a 0.2\% improvement over DeiT-B. This performance boost comes with a slight increase in computational cost, reflected in the higher FLOPs. Similar to the detection results, KAT shows more significant gains in smaller models. However, it still falls short compared to models with hierarchical architectures, such as ConvNeXt, which benefit from more efficient structural design.


\subsection{Ablation Study and Analysis}

\textbf{Activation Function.} As GR-KAN can be considered as a special kind of MLP with group rational function, we do an ablation study to consider different types of activation for MLP and compare with our GR-KAN. Superficially, we replace the activation function in MLP in ViT-Ti/16 to different kinds, including GELU~\cite{hendrycks2016gaussian}, ReLU~\cite{fukushima1969visual}, SiLU~\cite{elfwing2018sigmoid}, PReLU~\cite{he2015delving} and PAU~\cite{Molina2020Padé}, and comparing them with KAT.

Table~\ref{tab:act_acc} summarizes the top-1 accuracy on the ImageNet-1k dataset for each activation function with ViT-Ti/16. The results indicate that traditional activation functions like ReLU and GELU perform similarly. Learnable activations like PReLU and PAU show an improvement. Notably, Our KAT-T achieves the highest accuracy at 74.6\%, outperforming GELU by 1.9\%. This suggests that GR-KAN, as used in KAT-T, can significantly enhance the expressiveness of MLPs in vision transformers.

In addition to accuracy, we analyzed the computational cost of different activations by measuring throughput and peak memory usage on an NVIDIA A5000 GPU (Table~\ref{tab:act_computation}). All activation functions showed similar peak memory usage. However, our method (KAT-T) showed slightly lower throughput compared to the baseline activations (e.g., ReLU, GELU, and SiLU), which are more efficient. This suggests that while KAT-T offers substantial accuracy improvements, there is a trade-off in computational efficiency, which may be attributed to the increased complexity of rational function computations.
\begin{table}[h]
    \caption{Throughput and Peak memory for different activation on A5000 GPU. Input size is fixed to $[64, 1000, 512]$.}
    \label{tab:act_computation}
    \centering
    \begin{tabular}{l|c|c|c|c|c}
    \toprule
    \textbf{Activation} & ReLU & GeLU & SiLU & PReLU &Ours  \\
    \midrule
    \textbf{Throughput} (batch/s) & 2654 & 2643 & 2668 &2644 & 2313 \\
    \textbf{Peak Memory} (M) & 1380 & 1380 & 1380 & 1380 & 1380\\
    \bottomrule
    \end{tabular}
\end{table}

\textbf{Benefit of CUDA Implementation.} To evaluate the efficiency improvements introduced by our CUDA implementation discussed in Section~\ref{sec:rat}, we conducted experiments to measure both forward pass speed and peak memory usage. Specifically, we compared our CUDA implementation against two alternative methods. The first is called \emph{Torch Looped}, which loops over each channel group, applies the rational function, and then concatenates the results. The second is called \emph{Torch Vectorized}. In this method, the input tensor is reshaped according to the channel groups, the rational function is applied in a vectorized manner, and the tensor is reshaped back to its original form. We compare these three implementation on A5000 GPU, under 1) different group number $g\in \{1,2,4,8,16\}$. 2) different input dim $D\in\{128, 256, 512, 1024,2048\}$
\begin{figure}[h]
    \centering
    \begin{subfigure}{0.45\linewidth}
        \centering
        \includegraphics[width=\linewidth]{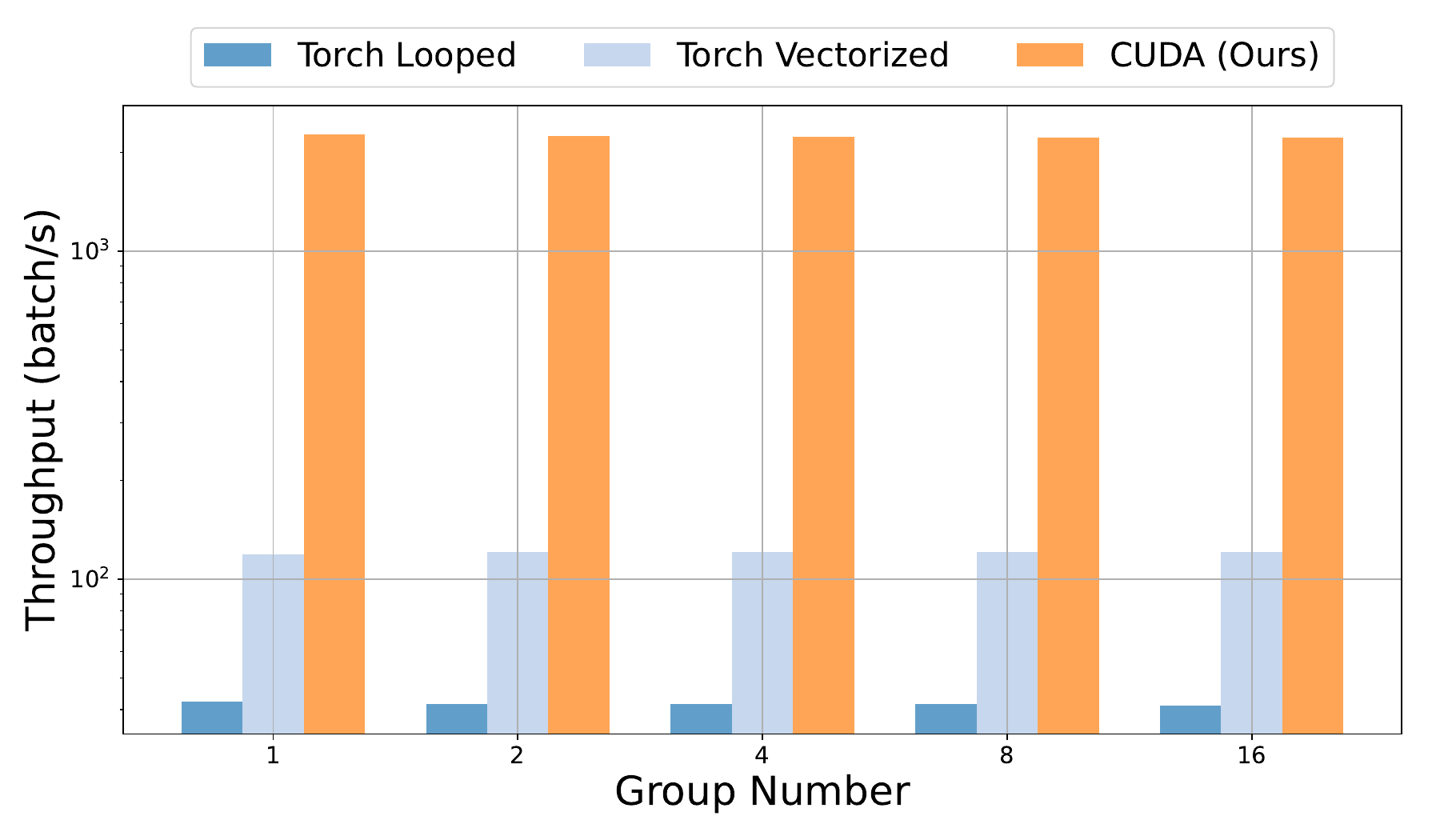}
        \caption{Throughput (batch/s) for Different Group Sizes. Larger the better.}
        \label{fig:throughput}
    \end{subfigure}
    \hfill
    \begin{subfigure}{0.45\linewidth}
        \centering
        \includegraphics[width=\linewidth]{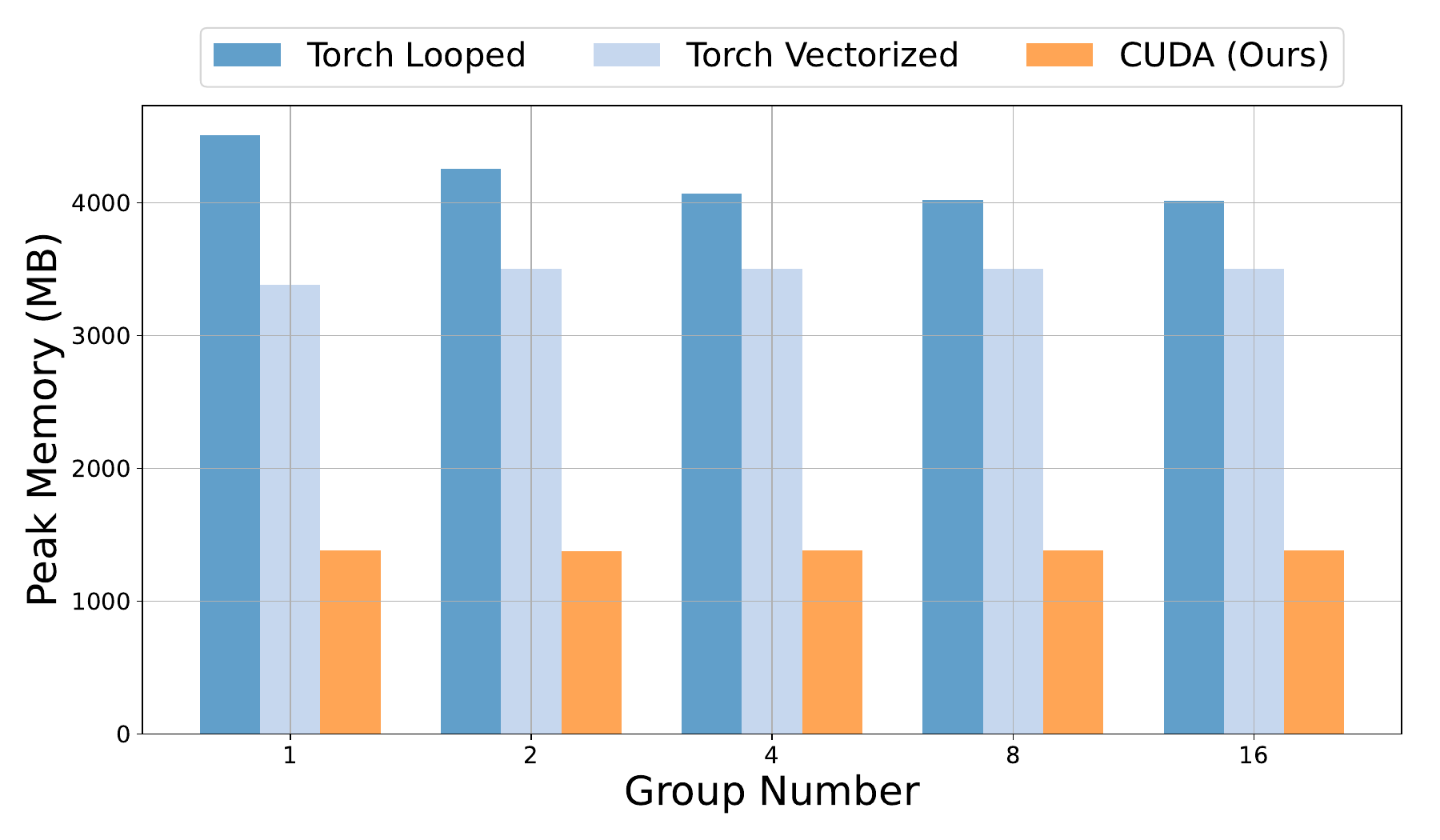}
        \caption{Peak Memory (MB) for Different Group Sizes. Smaller the better.}
        \label{fig:memory}
    \end{subfigure}
    \caption{Comparison of Throughput and Peak Memory for Different Methods and Group Sizes.  Input size is fixed to $[64, 1000, 512]$.}
    \label{fig:comparison_group}
\end{figure}
\begin{figure}[h]
    \centering
    \begin{subfigure}{0.45\linewidth}
        \centering
        \includegraphics[width=\linewidth]{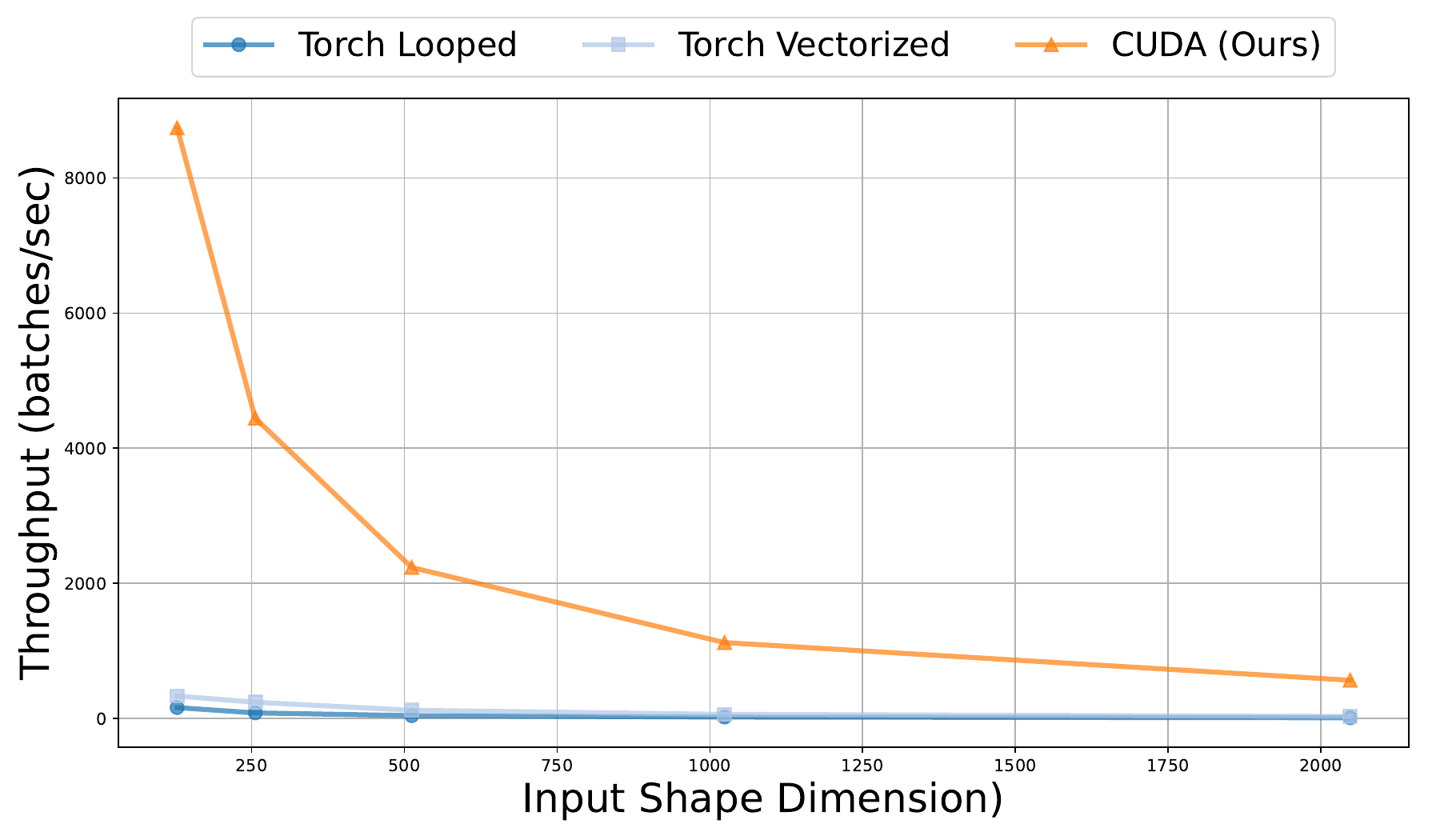}
        \caption{Throughput (batch/s) for Input Dimension Sizes. Larger the better.}
        \label{fig:throughput}
    \end{subfigure}
    \hfill
    \begin{subfigure}{0.45\linewidth}
        \centering
        \includegraphics[width=\linewidth]{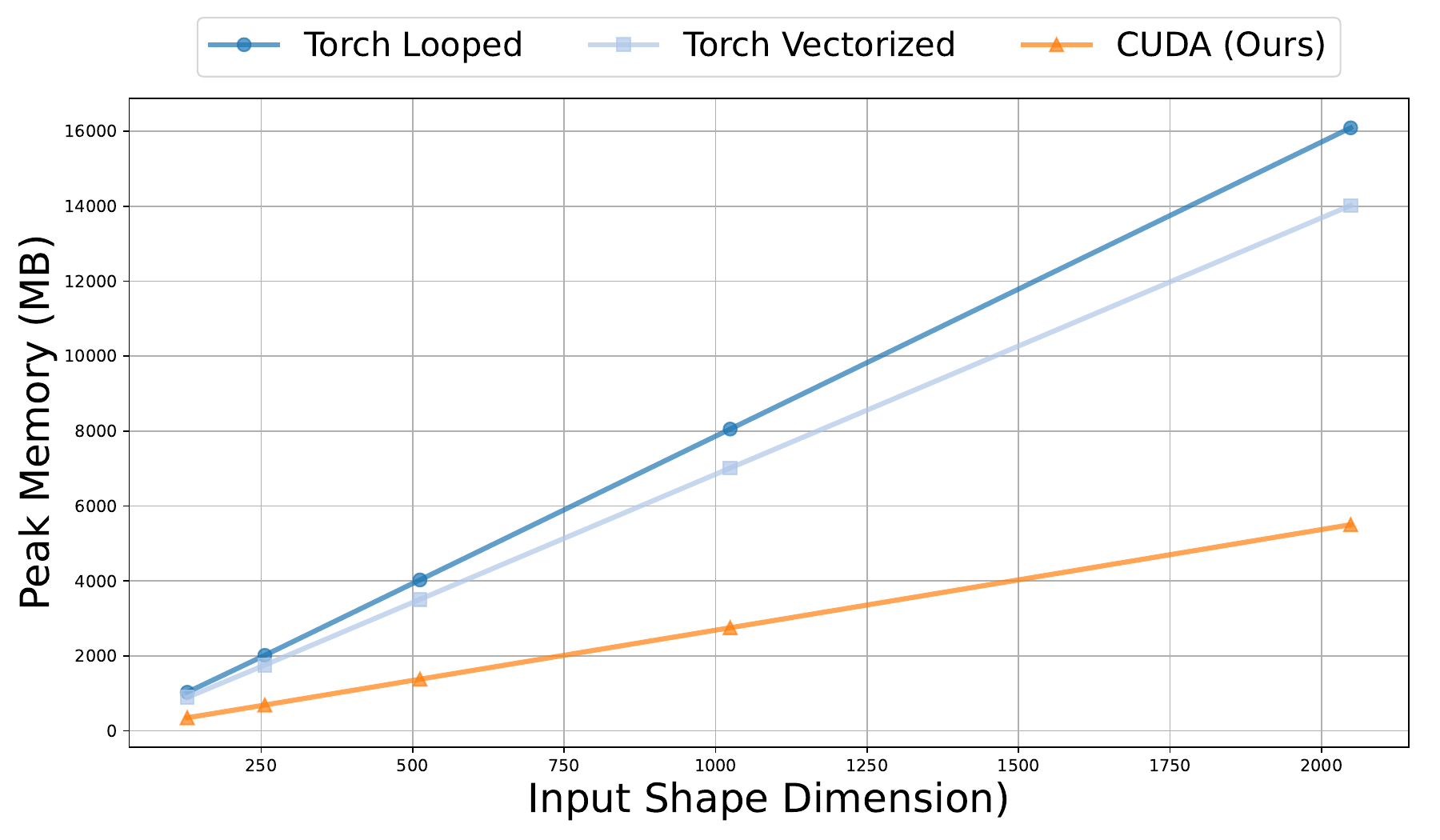}
        \caption{Peak Memory (MB) for Input Dimension Sizes. Smaller the better.}
        \label{fig:memory}
    \end{subfigure}
    \caption{Comparison of Throughput and Peak Memory for Different Methods and Input Dimension Sizes. Group size is fixed to 8.}
    \label{fig:comparison_D}
\end{figure}

The results, presented in Figure~\ref{fig:comparison_group} and Figure~\ref{fig:comparison_D}, clearly demonstrate that our CUDA implementation significantly outperforms both the Torch Looped and Torch Vectorized implementations, offering superior speed and memory efficiency.

\textbf{Rational Initialization.} We tested our KAT-T model with different initializations of the rational functions when training from scratch. As shown in Table~\ref{tab:rat-init}, the ``Identity - Swish'' initialization achieves the best performance, which we have adopted as our default setting.
\begin{table}[h]
    \centering
        \caption{Ablation on rational function initialization, with KAT-T.}
    \label{tab:rat-init}
    \begin{tabular}{c|c|c}
    \toprule
        \textbf{Rational 1 Init.} & \textbf{Rational 2 Init.} & \textbf{IN-1k Top-1} \\
        \midrule
        Identity & Identity & 69.7\\
        Swish & Swish & 74.4\\
        Identity & GeLU & 74.5\\
        \rowcolor{mygray}Identity & Swish & \textbf{74.6}\\
        \bottomrule
    \end{tabular}
\end{table}
       

\textbf{Visualization of Trained Functions}

\begin{figure}
    \centering
    \includegraphics[width=0.72\linewidth]{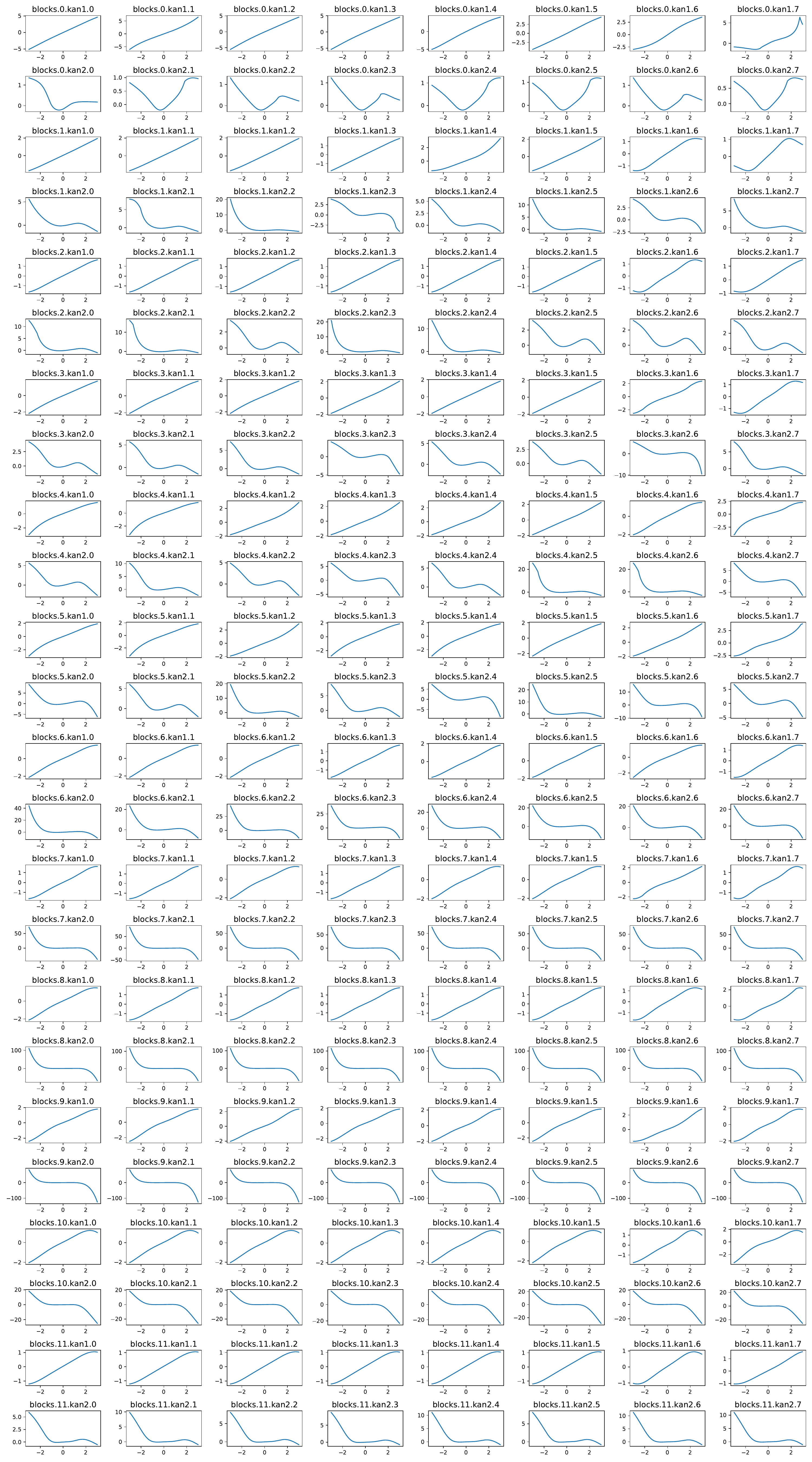}
    \caption{Fitted rational functions for KAT-S model, with 12 layers and 8 groups.}
    \label{fig:functions}
\end{figure}

An important aspect to examine is the behavior of the trained rational functions. As shown in Figure~\ref{fig:functions}, we plot the functions for KAT-S with $g=8$ across all 12 layers. The results indicate that within each layer, the rational functions exhibit similar trends, while the functions across different layers tend to differ from one another.

\section{Conclusion and Future Work}

In this work, we introduced the Kolmogorov–Arnold Transformer (KAT), a novel architecture that successfully integrates Kolmogorov-Arnold Networks (KANs) into transformers, addressing key challenges associated with large-scale training scenarios. Our proposed Group-Rational KAN (GR-KAN) variant, with its rational activation functions, group-based parameter sharing, and variance-preserving initialization, demonstrated significant improvements in computational efficiency and scalability. Through extensive experiments on vision tasks, including image recognition, object detection, and semantic segmentation, KAT outperformed traditional MLP-based transformers, achieving superior accuracy on ImageNet1K while maintaining comparable computational demands.

\textbf{Discussion.}
Our study highlights KAT's potential as a good alternative to MLP-based transformers, especially in large-scale vision tasks. This integration introduces exciting opportunities for broad applications. For example, employing KAT architectures might help development of language models.

However, KAT is not without its challenges. A primary concern is running speed. Even with the CUDA optimized code, the rational function is still slower than plain activation like ReLU and GELU. Another issue is the stability when using rational functions in neural networks. The higher order gradients for $a_m$ and $b_n$ can become unstable because of their dependence on the input power. Integrating these functions into the backpropagation process could introduce complications.

Additionally, it is important to acknowledge that our GR-KAN represents a hybrid model. On the one hand, GR-KAN is a KAN layer with shared edges and a rational base function. On the other hand, it can be interpret as MLP with a redesigned activation placed before the linear layer. It leverages the computational simplicity of MLPs but maintains some characteristics of KANs. However, GR-KAN is not a pure KAN model. Instead, it merges advantages from both systems to enhance overall functionality.

\textbf{Future Work.}
There are multiple directions of KAT for future research. One potential area of exploration is to find alternative base functions to further improve computational efficiency and compatibility with emerging hardware architectures. Currently, rational functions serve as one option, but other possibilities exist. These include Fourier transformations~\cite{fourierkan2024}, Wavelet transforms~\cite{bozorgasl2024wavkan}, and Gaussian radial bases~\cite{Li2024KolmogorovArnoldNA}.

Additionally, expanding the applicability of KAT to other domains beyond vision tasks, such as natural language processing or reinforcement learning, could unlock new opportunities for performance gains. Further research could also investigate hybrid models~\cite{yang2022deep,yu2023metaformer}, or adaptive mechanisms for dynamically selecting between KAN and MLP layers based on the complexity of the task, thereby optimizing resource utilization. Finally, addressing the remaining scalability challenges, particularly in terms of memory footprint and inference speed, will be crucial for deploying KAT in real-world applications at scale.

\section*{Acknowledgement}
We would like to acknowledge that computational work involved in this research work is partially supported by NUS IT’s Research Computing group using grant numbers
NUSREC-HPC-00001. We thank Weihao Yu, Qiuhong Shen and Runpeng yu for valuable discussions.


\bibliographystyle{alpha}
\bibliography{references}  

\section{Derivation and Calculation of FLOPs}

Given the function:
\[
F(x) = \frac{a_0 + a_1x + \dots + a_m x^m}{1 + |b_1x + \dots + b_n x^n|}
\]

\subsection{Plain Computation}

\subsubsection*{Numerator}
The numerator is a polynomial of degree \( m \):
\begin{itemize}
    \item Multiplications: There are \( \frac{m(m+1)}{2} \) multiplications for computing powers of \( x \) and \( m \) multiplications for coefficients \( a_i \), giving \( \frac{m(m+1)}{2} +m \).
    \item Additions: There are \( m \) additions to sum up the polynomial terms.
\end{itemize}

\subsubsection*{Denominator}
The denominator involves the absolute value of a polynomial of degree \( n \):
\begin{itemize}
    \item Multiplications: There are \( \frac{n(n+1)}{2} \) multiplications for powers of \( x \) and \( n \) multiplications for coefficients \( b_i \), giving \( \frac{n(n+1)}{2} + n \).
    \item Additions: There are \( n \) additions for polynomial terms and 1 additional addition after the absolute value operation.
    \item Absolute value operation: 1 absolute value calculation.
\end{itemize}

\textbf{Division.} There is 1 division operation for the final computation of \( F(x) \).

\textbf{Total FLOPs.}
The total FLOPs for any \( m \) and \( n \) are:
\[
\text{Multiplications: } \frac{m(m+1)}{2} + \frac{n(n+1)}{2} + m + n + 1, \quad \text{Additions: } m + n + 1, \quad \text{Absolute Value: } 1, \quad \text{Division: } 1
\]

In case \( m = 5 \) and \( n = 4 \), there are totally 34 multiplications, 10 summations, 1 absolute value and 1 division. In total 46.

\subsection{Horner's Method}

Using horner's method, for a polynomial of order $m$, we need $m$ summations and $m$ multiplications. 

Thus, for numerator, we need $m$ summations and $m$ multiplications. For denominator, we need $n+1$ summations and $n$ multiplications. In total, we need $m+n+1$ summation, $m+n$ multiplications, $1$ absolute value, and $1$ division.

In case \( m = 5 \) and \( n = 4 \), there are a total of 21 FLOPs, comprising 9 multiplications, 10 summations, 1 absolute value, and 1 division.

\section{Hyper-parameters for KAT model}
The hyper-parameter for training KAT model on ImageNet-1k is shown in Table~\ref{tab:hyper}.
\begin{table}[ht]
\centering
\caption{Hyper-parameters of KAT on ImageNet image classification.}
\label{tab:hyper}
\begin{tabular}{lccc}
\toprule
\textbf{} & \multicolumn{3}{c}{\textbf{KAT}} \\
\cmidrule(lr){2-4}
\textbf{} & \textbf{Tiny} & \textbf{Small} & \textbf{Base} \\
\midrule
Input resolution & \multicolumn{3}{c}{$224^2$} \\
Epochs & \multicolumn{3}{c}{300} \\
Batch size & \multicolumn{3}{c}{1024} \\
Optimizer & \multicolumn{3}{c}{AdamW} \\
Adam $\epsilon$ & \multicolumn{3}{c}{$1 \times 10^{-8}$} \\
Adam ($\beta_1$, $\beta_2$) & \multicolumn{3}{c}{(0.9, 0.999)} \\
Learning rate & \multicolumn{3}{c}{$1 \times 10^{-3}$} \\
Learning rate decay & \multicolumn{3}{c}{Cosine} \\
Gradient clipping & \multicolumn{3}{c}{None} \\
Warmup epochs &  \multicolumn{3}{c}{5} \\
Weight decay & \multicolumn{3}{c}{0.05} \\
Rand Augment & \multicolumn{3}{c}{9/0.5} \\
Repeated Augmentation & \multicolumn{3}{c}{off} \\
Cutmix & \multicolumn{3}{c}{1.0} \\
Mixup & \multicolumn{3}{c}{0.8} \\
Cutmix-Mixup switch prob & \multicolumn{3}{c}{0.5} \\
Random erasing prob & \multicolumn{3}{c}{0.25} \\
Label smoothing & \multicolumn{3}{c}{0.1} \\
Peak stochastic depth rate & 0.1 & 0.1 & 0.4  \\
Random erasing prob & \multicolumn{3}{c}{0.25} \\
EMA decay rate & \multicolumn{3}{c}{0.9999} \\
\bottomrule
\end{tabular}
\end{table}

\end{document}